\documentclass{ecai} 

\usepackage{latexsym}
\usepackage{amssymb}
\usepackage{amsmath}
\usepackage{amsthm}
\usepackage{booktabs}
\usepackage{enumitem}
\usepackage{graphicx}
\usepackage{color}
\usepackage{multirow}
\usepackage{array}

\newcommand{\BibTeX}{B\kern-.05em{\sc i\kern-.025em b}\kern-.08em\TeX}

\begin{document}

\nolinenumbers
\begin{frontmatter}


\paperid{2139} 


\title{Advancing Topic Segmentation of Broadcasted Speech with Multilingual Semantic Embeddings}


\author[A]{\fnms{Sakshi Deo}~\snm{Shukla}\thanks{Corresponding Author. Email: sakshishukla1996@gmail.com.}}
\author[B]{\fnms{Pavel}~\snm{Denisov}}
\author[B]{\fnms{Tugtekin}~\snm{Turan}} 

\address[A]{Fraunhofer Institute for Digital Media Technology, Germany}
\address[B]{Fraunhofer Institute for Intelligent Analysis and Information Systems, Germany}


\begin{abstract}
Recent advancements in speech-based topic segmentation have highlighted the potential of pretrained speech encoders to capture semantic representations directly from speech. Traditionally, topic segmentation has relied on a pipeline approach in which transcripts of the automatic speech recognition systems are generated, followed by text-based segmentation algorithms. In this paper, we introduce an end-to-end scheme that bypasses this conventional two-step process by directly employing semantic speech encoders for segmentation. Focused on the broadcasted news domain, which poses unique challenges due to the diversity of speakers and topics within single recordings, we address the challenge of accessing topic change points efficiently in an end-to-end manner. Furthermore, we propose a new benchmark for spoken news topic segmentation by utilizing a dataset featuring approximately 1000 hours of publicly available recordings across six European languages and including an evaluation set in Hindi to test the model's cross-domain performance in a cross-lingual, zero-shot scenario. This setup reflects real-world diversity and the need for models adapting to various linguistic settings. Our results demonstrate that while the traditional pipeline approach achieves a state-of-the-art $P_k$ score of 0.2431 for English, our end-to-end model delivers a competitive $P_k$ score of 0.2564. When trained multilingually, these scores further improve to 0.1988 and 0.2370, respectively. To support further research, we release our model along with data preparation scripts, facilitating open research on multilingual spoken news topic segmentation. 


\end{abstract}

\end{frontmatter}


\section{Introduction}

The explosive emergence of multimedia in recent years has initiated a paradigm shift in traditional content consumption methods. News media, a crucial component of mass communication, occupies a central role as a widely utilized and readily accessible source of information. The popularity of podcasts, online radio, and streaming platforms has highlighted the demand for topic segmentation \cite{chen2017modeling}. This shift has necessitated the development of sophisticated methods for structuring and processing large streams, allowing users to access news content anytime and anywhere, thus emphasizing the critical nature of efficient content management \cite{harrando2021and}.

Topic segmentation, dividing long texts or audio streams into topically coherent units, is an important task in natural language processing (NLP) that simplifies downstream applications such as summarization, information retrieval, and personalized content delivery. While this task has been widely explored in the text domain \cite{badjatiya2018}, its application to speech recordings often relies heavily on prosodic information rather than semantic cues \cite{berlage2020}. More advanced models can leverage automatic speech recognition (ASR) and dense text representations produced by pretrained neural networks, significantly enhancing the capability to segment multimedia documents into distinct topic units \cite{ghinassi2021}. 
Despite these advancements, the segmentation of spoken audio presents unique challenges. Traditional methods typically rely on ASR to convert speech to text and segment using various NLP techniques \cite{sehikh2017rnn}. This solution may introduce errors during the transcription process, which complicates the segmentation and potentially leads to inaccuracies in topic identification. Moreover, the contextual representation of spoken language, which includes intonation, pauses, and other speech nuances, remains weakly captured by current segmentation methods, thus limiting the effectiveness of topic identification in spoken dialogues.

 Recognizing the limitations, we aim to propose a comparative analysis of two innovative architectures to improve topic segmentation in spoken audio. The first one, a \textit{pipeline} approach, utilizes the Whisper model \cite{radford2023robust} for ASR, followed by segmentation using neural sentence encoders. This approach, while robust, is contingent on the accuracy of the transcription process and necessitates extensive manual intervention. Conversely, our proposed \textit{end-to-end} method seeks to circumvent the transcription phase altogether by directly extracting semantic latent representations from the audio using the novel Sentence-Level Multimodal and Language-Agnostic Representations (SONAR) encoder \cite{duquenne2023sentence}. Our model SpeechTopSeg\footnote{\url{https://github.com/sakshishukla1996/SpeechTopSeg}} is available with the data preparation scripts and checkpoints \cite{shukla_2024_13338560}.

SONAR consists of speech and sentence encoders with a common sentence decoder sharing common semantic latent representations for sentence and speech which allows a thorough comparative study with the proposed approaches. Moreover, SONAR is multilingual therefore, language-specific encoders can be utilized for various analyses. Since the end-to-end approach capitalizes on recent breakthroughs in neural architectures that encode both the topical and structural information of spoken content, thereby predicting topic shifts directly from the audio stream without the intermediate transcription steps and highlighting the potential of the end-to-end model compared to the pipeline, for reducing the model complexity in the number of parameters.

In this paper, we aim to refine how spoken audio is segmented into topics and redefine the process by introducing more efficient, direct methods. We evaluate the effectiveness of our proposed approaches against established benchmarks in the field, showcasing their potential to change how we interact with multimedia content. Apart from our analysis, we introduce a new speech segmentation dataset based on YouTube content. This dataset includes publicly available news content across six European languages, additionally featuring an evaluation set in Hindi as a zero-shot scenario. The proposed dataset serves as a key tool in training and evaluations of our systems, helping to bridge the gap between theoretical research and practical applications. By utilizing real-world news data, we can understand the complexities of spoken language and the effectiveness of a segmentation model efficiently. Our results indicate that the end-to-end model delivers competitive scores under a multilingual setup.

The rest of this paper is structured as follows: we introduce related work in Section 2. Next, we describe the task and dataset creation process in Section 3. We formalize our methods in Section 4. We report results and insights from the evaluation in Section 5 and Section 6, respectively. Finally, we conclude in Section 7 with future work.

\section{Related Work}
\subsection{Topic Segmentation Systems}
Traditionally, topic segmentation in text has primarily revolved around comprehending lexical representations and extracting semantic meanings through unsupervised learning methodologies. For instance, \citet{chien2011topic} utilized text streams and applied Latent Dirichlet Allocation (LDA) on bag-of-words to segment topics. Text sequences are treated as blocks, and topic distribution probabilities are computed to identify coherently similar blocks. Similarly, \citet{riedl2012topictiling} adapted the concept of Text Tiling to topic segmentation by using topic IDs instead of bag-of-words, employing the LDA algorithm to predict topic changes.

The advent of deep learning approaches has introduced benchmark models for text segmentation. For instance, \citet{li2018segbot} introduced a pointer network comprising three components to detect overall topic boundaries from various possibilities. Meanwhile, \citet{koshorek2018text} framed topic segmentation as a supervised learning task, employing an architecture comprising two Bidirectional Long Short-Term Memory (BLSTM) networks \cite{hochreiter1997long,graves2005bidirectional}. The lower-level network generates sentence embeddings, which are fed into the higher level to predict topic boundaries.

Recently, speech-based topic segmentation has gained traction with the development of speech embedding generators. \citet{berlage2020} introduced a novel task of utilizing audio embeddings for automated, topically coherent segmentation of radio shows, employing three different audio embedding generators for effective segmentation. Additionally, \citet{Ghinassiaudio} extended the implementation of topic segmentation in the speech by solely utilizing pretrained neural audio embeddings without further finetuning the audio encoders. Various pretrained speech encoders were combined with segmentation models to predict topic boundaries using fixed input window lengths of 1-second non-overlapping windows. The results confirm that neural representations can improve topic segmentation over non-neural features, particularly in specific domains.

\subsection{Topic Segmentation Datasets}
The most common benchmark datasets for text-based topic segmentation are Wiki727k \cite{koshorek2018text}, Wiki50 \cite{arnold2019sector}, and Choi dataset \cite{choi2000advances}. However, only a few open-source datasets are available for speech topic segmentation, all in the English language. The AMI \cite{mccowan2005ami} and ICSI~\cite{janin2003icsi} are recorded meeting corpora, which can be used for various multimodal tasks, including speech topic segmentation. The TDT challenges \cite{cieri1999tdt} publish a couple more datasets for speech topic segmentation, but they are available for a fee only. Addressing the data availability issue, we propose a new combined dataset and release its preparation steps for future research.


\section{Dataset}
\label{section:data}
\subsection{Multilingual Topic Segmentation Dataset}
\label{section:data:multilingual}
We construct a multilingual topic segmentation dataset based
on the "Latest news bulletin" program from the Euronews channel.
We aim to construct a dataset that can be used for real-world applications, therefore there are always speaker overlaps for recordings from the same TV station and no speaker overlaps between different TV stations, our dataset includes both conditions. Also, there are almost always multiple languages in the reports on international events because of code-switching. Though, this poses a challenge for a monolingual speech encoder, news broadcasts without this condition are too rare to consider. The Euronews program proves to be no other exception to these challenges.
Video recordings of this program are publicly available
on YouTube in multiple languages, grouped in playlists
by language\footnote{For example, videos in English:
\url{https://www.youtube.com/playlist?list=PLSyY1udCyYqA0klB0MeTnE4cKTqX_MnAi}}.
Some videos are annotated with chapter timestamps according to
news topics covered in a video. We request metadata for each video in a playlist
and download the audio part if the metadata contains chapter annotation.
The resulting collections of recordings are split into the
training, validation, and evaluation sets roughly in 90/5/5\% proportions for each language.
This source of recordings makes it challenging to avoid speaker
overlap between the splits within one language,
and we do not attempt it. However, we sort the recordings
chronologically before splitting them to minimize
topic and vocabulary overlaps between the sets.
All recordings are downloaded and encoded in Opus, in 48 kHz sampling rate, and with the variable bitrate between 80 and 180 kbits/s per recording (on average 107 kbits/s overall for all recordings)
and is subsequently converted to the single-channel WAV format
with a 16 kHz sampling rate.
All videos were accessed on March 3, 2024.
After checking the playlists in 9 languages, we found that the playlists with
the videos in Russian, Bulgarian, and Romanian languages did not contain any
videos with chapter annotation, the other 6 languages could be included in the dataset. 
The detailed statistics of the complete Euronews dataset can be seen in Table~\ref{tab:data:euronews}.

\begin{table}[h]
\centering
\setlength{\tabcolsep}{0.7mm}
\begin{tabular}{l rrr rrr rrr}
\toprule
\multirow{2}{1.5cm}{\emph{Source} Language} & \multicolumn{3}{c}{Total duration (hrs)} & \multicolumn{3}{c}{\#recordings} & \multicolumn{3}{c}{avg(\#seg$/$rec)} \\
\cmidrule{2-10}
& \multicolumn{1}{c}{train} & \multicolumn{1}{c}{dev} & \multicolumn{1}{c}{test} & \multicolumn{1}{c}{train} & \multicolumn{1}{c}{dev} & \multicolumn{1}{c}{test} & \multicolumn{1}{c}{train} & \multicolumn{1}{c}{dev} & \multicolumn{1}{c}{test} \\
\midrule
\emph{Euronews} & & & & & & & & \\
English & 226.31 & 10.05 & 11.16 & 1168 & 65 & 65 & 7.11 & 7.33 & 7.89 \\
French & 191.14 & 10.98 & 11.56 & 1204 & 67 & 67 & 6.89 & 7.10 & 7.71 \\
German & 154.23 & 8.46 & 9.94 & 980 & 54 & 55 & 7.28 & 7.53 & 7.85 \\
Italian & 135.18 & 7.85 & 8.78 & 861 & 48 & 48 & 6.95 & 6.68 & 8.16 \\
Portuguese & 46.43 & 2.14 & 2.24 & 286 & 16 & 16 & 7.05 & 6.68 & 6.75 \\
Spanish & 150.74 & 10.03 & 9.71 & 465 & 26 & 26 & 10.88 & 15.88 & 11.69 \\
Total & 904.03 & 49.51 & 53.39 & 4964 & 276 & 277 & 7.69 & 8.53 & 8.34 \\
\midrule
\emph{Tagesschau} & & & & & & & & \\
German & 217.09 & 16.17 & 12.13 & 837 & 45 & 45 & 10.60 & 10.08 & 10.08 \\
\midrule
\emph{Akashvani} & & & & & & & & \\
Hindi &- &- & 14.96 &- &- & 51 & - & - & 11.98  \\
\bottomrule
\end{tabular}
\caption{Complete details of the dataset from Euronews, Tagesschau and Akashvani news media platforms.}
\label{tab:data:euronews}
\end{table}


\subsection{Cross Domain Topic Segmentation Dataset}

We curated a cross-domain topic segmentation dataset in German, utilizing recordings from the "News Bulletin at 8 pm" program aired on Tagesschau. These recordings complement the German portion of the Euronews dataset, as they come from another television station, and can be used to evaluate the model's robustness and generalization across different audio effects and narration styles, which we regard as different domains. The video recordings are publicly available on YouTube\footnote{German video: \url{https://www.youtube.com/watch?v=-kEMDZFiHMs}}.
To ensure consistent metadata preparation with the Euronews dataset, we applied identical preprocessing steps to the Tagesschau dataset. The videos were accessed on 20 January 2024, and the detailed statistics of the dataset can be seen in Table \ref{tab:data:euronews}.



\subsection{Zero-Shot Topic Segmentation Dataset}
We curated an evaluation set for the task of topic segmentation in a novel language, Hindi, to assess the cross-domain performance of the model in a cross-lingual and zero-shot scenario. We utilized radio-broadcasted recordings from News on Akashvani All India Radio (AIR), available on Youtube\footnote{Hindi audio: \url{https://www.youtube.com/watch?v=NakBvRE0V4o}} channel, organized into the Hindi 2.0 playlist. The playlist includes bulletin news broadcasts aired four times a day, spanning morning, afternoon, evening, and late-night bulletins. Each broadcast exhibits significant variation from the others, reflecting the evolving events and updates throughout the day. Since the audio recordings lack predefined chapters indicating changes in the topic, we manually annotate the chapters in the recordings with the assistance of a native Hindi speaker. To maintain consistency in metadata preparation with the other two datasets, we curated the evaluation set in a single-channel WAV format with a sampling rate of 16kHz. The detailed statistics of the dataset can be seen in Table \ref{tab:data:euronews}.




\section{Methods}

\subsection{Pipeline}

The conventional pipeline approach relies on the two-step process.
The first step utilizes an ASR system, possibly combined with a speaker diarization system,
and converts speech input $\mathbf{x}^{S} = \{ x^{S}_{1}, \dots, x^{S}_{n^{S}} \}$
to a text representation $\mathbf{x}^{T}= \{ x^{T}_{1}, \dots, x^{T}_{n^{T}} \}$,
also named a \emph{transcription}, where $n^{S} \in \mathbb{Z}$
and $n^{T} \in \mathbb{Z}$ are numbers of samples in the speech and
tokens in the transcription, $x^{S}_{i} \in \mathbb{Z}$ is a quantized amplitude value of
the recording sample, $x^{T}_{j} \in \mathbb{N}: 1 \le x^{T}_{j} \le v$
is an index of token in a vocabulary, and $v \in \mathbb{N}$ is the vocabulary size.
The second step consumes the transcription as a sequence of blocks
$\mathbf{X}^{B} = \{ \mathbf{s}_{1}, \dots, \mathbf{s}_{n^{B}} \}$
of a length $n^{B} \in \mathbb{Z}$,
where each block $ \mathbf{s}_{k} = \{ x^{T}_{1 + n^{s}_{k-1}}, \dots, x^{T}_{n^{s}_{k} + n^{s}_{k-1}}  \} $
is a sequence of token indices in one block, and applies a transcription
segmentation algorithm to this sequence of blocks.
Length of a $k$-th transcription block $n^{s}_{k} \in \mathbb{Z}$ can be a constant number,
or can be defined by the speaker diarization or turn detection algorithms integrated with the ASR system.
Output of a transcription segmentation algorithm
is a sequence of binary labels $\mathbf{y} = \{y_{1}, \dots, y_{n^{B}}\}$, 
where each label $y_{k} \in \{0, 1\}$
denotes whether the respective block $\mathbf{s}_{k}$ belongs to the same topic
as the next block $\mathbf{s}_{k+1}$.



\textbf{Non-neural baseline.} Our non-neural baseline is a well-known text-based topic
segmentation algorithm called TextTilling \cite{hearst-1997-text}, as implemented
by the Natural Language Toolkit (NLTK) \cite{bird2009natural}. Moreover, we include the vanilla TextTilling approach as our approach is based on supervised learning. 
TextTilling operates on sequences of tokens of a fixed length,
originally referred to as token sequences.
These token sequences are organized into blocks of token sequences
of a fixed length as well, and the similarity score is calculated
for a gap $g$ between each pair of neighboring blocks number $g$ and $g+1$:
\begin{equation*}
\text{similarity}(g) = \frac{\sum_{t} w_{t,g} w_{t,g+1}}{\sqrt{\sum_{t} w_{t,g}^2 \sum_{t} w_{t,g+1}^2}},
\end{equation*}
where $1 \le t \le n^{TT}$, $n^{TT} \in \mathbb{Z}$ is a number of tokens in a block of token-sequences,
and $w_{a,b}$ denotes a number of occurrences of the token number $a$ of the block number $b$
within that block.
The sequence of similarity scores is then processed with a moving average smoothing: score
value for each gap is replaced with its average and a certain number of values before and after it.
Finally, the depth score values are obtained from the sequence of similarity scores:
\begin{equation*}
\text{depth}(g) = \text{similarity}(l) + \text{similarity}(r) - 2\text{similarity}(g),
\end{equation*}
where $l$ and $r$ denote indices of gaps corresponding
to the local maximums in the sequence of similarity scores
to the left and the right of the gap $g$. 
Gaps with higher depth scores correspond
to pairs of blocks with low mutual similarity compared
to pairs of blocks before and after them,
which can point to a change of topic.
TextTilling assigns a topic change label to the gaps with the depth score
above $\overline{s} - \sigma / 2$, where $\overline{s}$ and $\sigma$ are
the average and standard deviation of all depth scores.
This method assumes that tokens correspond to words.


\textbf{Neural baseline.} We adopt the BLSTM-based text segmentation model, TextSeg, proposed by \citet{koshorek2018text}
as our neural baseline. Each block of tokens $\mathbf{s}_{k}$ is passed through a token embedding
layer, a BLSTM network, and a maximum pooling operation to obtain a single vector
embedding of the block:
\begin{align*}
\mathbf{H}^{\bf Emb}_{k} &= \text{Embedding}(\mathbf{s}_{k}) \\
\mathbf{H}^{\bf Enc}_{k} &= \text{BLSTM}^{\bf Enc}(\mathbf{H}^{\bf Emb}_{k}) \\
\mathbf{z}_{k} &= \text{MaxPool}(\mathbf{H}^{\bf Enc}_{k}),
\end{align*}
where $\mathbf{H}^{\bf Emb}_{k} \in \mathbb{R}^{n^{s}_{k} \times d}$
and $\mathbf{H}^{\bf Enc}_{k} \in \mathbb{R}^{n^{s}_{k} \times d}$ are hidden representations
of tokens in the block $k$, $\mathbf{z}_{k} \in \mathbb{R}^{d}$, and $d$ is the hidden size.
The complete sequence of block embeddings
$\mathbf{Z} = \{ \mathbf{z}_{1}, \dots, \mathbf{z}_{n^{B}} \}$ 
is then processed by another BLSTM network, and is projected
to a 2-dimensional output representation through a linear layer:
\begin{align*}
\mathbf{H}^{\bf Head} &= \text{BLSTM}^{\bf Head}(\mathbf{Z}) \\
\mathbf{O} &= \text{Linear}(\mathbf{H}^{\bf Head}),
\end{align*}
where $\mathbf{H}^{\bf Head} \in \mathbb{R}^{n^{B} \times d}$,
and $\mathbf{O} = \{ \mathbf{o}_1, \dots, \mathbf{o}_{n^{B}} \}$ is a sequence of output
representations of blocks $\mathbf{o}_k \in \mathbb{R}^{2}$.
Finally, the softmax function is applied to the output block representation,
yielding a two-element vector with probabilities
of topic change and lack of such.

This method assumes that tokens correspond to words.
The word embeddings are pretrained and kept unchanged
during model training, while all other parameters are trained from scratch.



\begin{figure}[!t]
    \centering
    \includegraphics[width=0.8\linewidth]{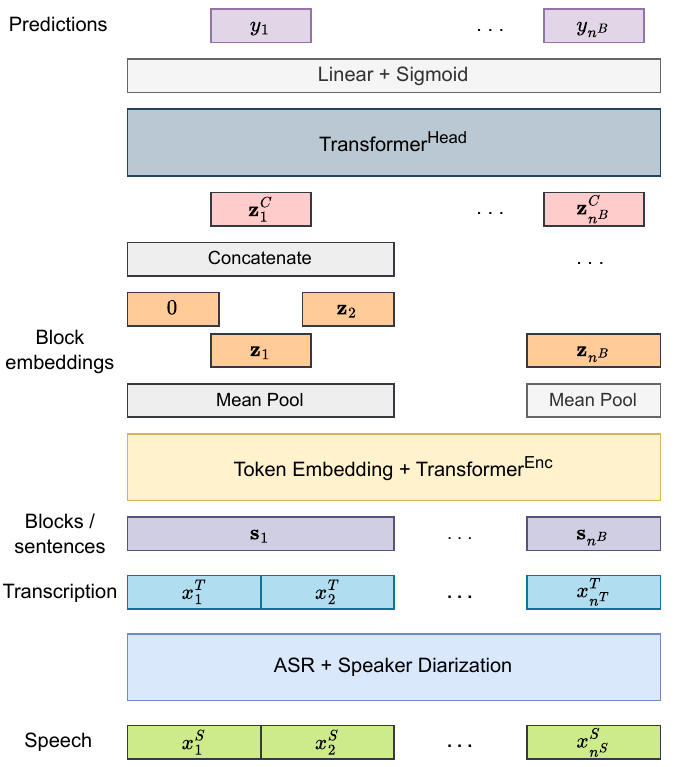}
    \caption{Our pipeline system.}
    \label{fig:pipeline}
\end{figure}

\textbf{Proposed method.} Our text segmentation model (Figure \ref{fig:pipeline}) introduces several modifications
compared to the neural baseline.
First, we replace BLSTM networks with Transformer \cite{vaswani2017attention}
and maximum pooling with mean pooling:
\begin{align*}
\mathbf{H}^{\bf Emb}_{k} &= \text{Embedding}(\mathbf{s}_{k}) \\
\mathbf{H}^{\bf Enc}_{k} &= \text{Transformer}^{\bf Enc}(\mathbf{H}^{\bf Emb}_{k}) \\
\mathbf{z}_{k} &= \text{MeanPool}(\mathbf{H}^{\bf Enc}_{k})
\end{align*}\\
Second, we extend each block embedding with the context of the previous and next blocks before
feeding it the segmentation network:
\begin{align*}
\mathbf{z}^{C}_{k} &= (\mathbf{z}_{k-1}, \mathbf{z}_{k}, \mathbf{z}_{k+1}) \\
\mathbf{H}^{\bf Head} &= \text{Transformer}^{\bf Head}(\mathbf{Z}^{C})
\end{align*}
where $\mathbf{Z}^{C} = \{ \mathbf{z}^{C}_{1}, \dots, \mathbf{z}^{C}_{n^{B}} \}$,
$\mathbf{z}^{C}_{k} \in \mathbb{R}^{3d}$,
and $\mathbf{z}_{k}$ is filled with zeros for $k=0$ and $k=n^{B}+1$.
Third, we have only one dimension in the output block representation
$\mathbf{o}_k$ and use the sigmoid function to map it to a value that is interpreted
as a probability of topic change.

This method assumes that tokens are subwords.
The subword tokenizer, token embeddings, and encoder Transformer
are pretrained and kept unchanged during the model training,
while the head Transformer and output layer are trained from scratch.


\subsection{End-to-End}

\begin{figure}[!t]
    \centering
    \includegraphics[width=0.75\linewidth]{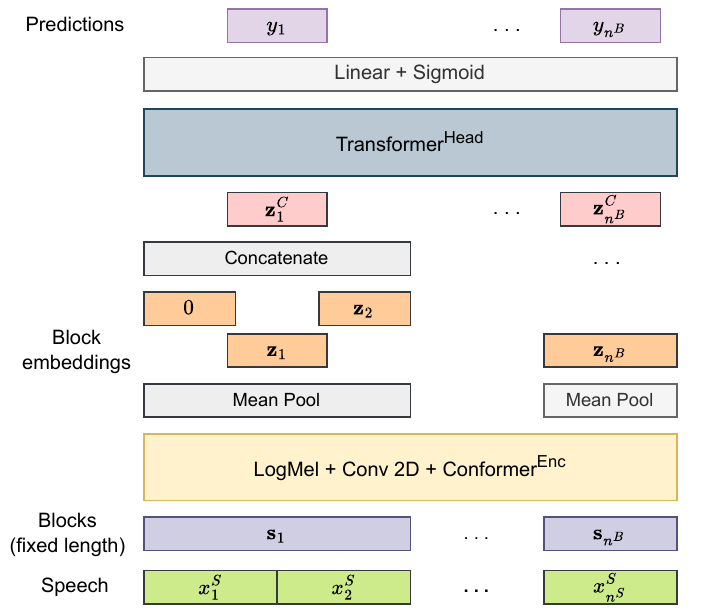}
    \caption{Our end-to-end system.}
    \label{fig:e2e}
\end{figure}

We utilize a pretrained encoder to obtain a fixed-size embedding of each block of tokens in the pipeline system.
Another pretrained encoder is available, which outputs a similar fixed-size embedding directly from the speech input.
This enables us to omit the ASR step of the pipeline system and to build an end-to-end speech topic segmentation model (Figure \ref{fig:e2e}).
We redefine the input blocks to contain sequences of audio samples instead of text tokens:
$ \mathbf{s}_{k} = \{ x^{S}_{1 + n^{s}_{k-1}}, \dots, x^{S}_{n^{s}_{k} + n^{s}_{k-1}}  \} $.
A block of audio samples is converted to a hidden representation
using a log-mel filter bank feature extractor followed by a
2D convolutional neural network and a linear layer.
The hidden representation is then processed by Conformer  \cite{gulati2020conformer}:
\begin{align*}
\mathbf{F}_{k} &= \text{LogMel}(\mathbf{s}_{k}) \\
\mathbf{H}^{\bf Emb}_{k} &= \text{Linear}^{\bf Enc}(\text{Conv2D}(\mathbf{F}_{k})) \\
\mathbf{H}^{\bf Enc}_{k} &= \text{Conformer}^{\bf Enc}(\mathbf{H}^{\bf Emb}_{k})
\end{align*}
The remaining part of the model is identical to our text-based topic segmentation model.
The convolutional neural network, introduced linear layer, and Conformer are pretrained
and kept unchanged during model training,
while the head Transformer and output layer are trained from scratch.

\section{Experimental setup}

\subsection{Metrics}

\textbf{\pmb{$P_k$} Score.} We utilize the $P_k$ metric, as defined by \citet{beeferman1999statistical}, to evaluate our systems. $P_k$ denotes the probability that, while traversing a sliding window of size $k$ across sentences, the sentences at the window's edges will be inaccurately classified as belonging to the same segment. The $P_k$ score is a penalty metric. It is calculated according to the following equation:

{
\footnotesize
\begin{align*}
p(&\text{error} | \mathtt{ref}, \mathtt{hyp}, k) = \\
& p(\text{miss}| \mathtt{ref}, \mathtt{hyp}, \text{different }\mathtt{ref}\text{ segments}, k) p(\text{different }\mathtt{ref}\text{ segments}| \mathtt{ref}, k) \\
& + p(\text{false alarm}| \mathtt{ref}, \mathtt{hyp}, \text{same }\mathtt{ref}\text{ segment}, k) p(\text{same }\mathtt{ref}\text{ segment} | \mathtt{ref}, k)
\end{align*}
}

\begin{figure}[htb]
    \centering
    \includegraphics[width=0.8\linewidth]{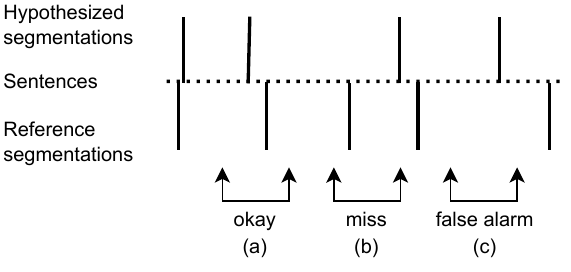}
    \caption{The (a) is an acceptable boundary, (b) represents \textit{false negative} where the true segment is present but is not predicted, and (c) represents a false alarm where the true segment doesnot exist but is predicted.}
    \label{fig:pk_segment_boundary}
\end{figure}

The lower the score, the greater the model's robustness. Figure \ref{fig:pk_segment_boundary} shows how exactly $P_k$ is calculated. As our baseline approach is based on \citet{koshorek2018text} model, we set our $k$ score to half of the average segment size in the ground-truth segmentation.

\textbf{Windiff Score.} While $P_k$ serves as the benchmark metric for topic segmentation, it presents several challenges, penalizing only false-negative. We employ another sliding window-based metric, Windiff \cite{pevzner2002critique}, to provide a more comprehensive understanding of the model's robustness. Windiff operates by moving a sliding window across the sentences and counting the instances where the hypothesized (predicted) and reference (ground-truth) segment boundaries differ within the window. The value of $k$ remains consistent with that used in $P_k$. We use SEGEVAL\footnote{\url{https://github.com/cfournie/segmentation.evaluation}} package \cite{fournier2013evaluating} to perform our evaluations.

{
\footnotesize
\begin{align*}
        \text{Windiff}(\text{ref}, \text{hyp}) = \frac{1}{N - k} \Sigma^{N-k}_{i=1} (|b(\text{ref}_i. \text{ref}_{i+k}) - b(\text{hyp}_i, \text{hyp}_{i+k})| > 0)
\end{align*}
}


\textbf{Segmentation Purity and Coverage F-score.}

$P_k$ and WinDiff depend a lot on the segment length of a particular system and do not suit well for comparisons across systems.
If the sliding window length of $P_k$ is larger than some missed reference segment and this segment happens to be inside of the sliding window, $P_k$ will not take it into account, because the window's start and end still belong to different segments, thus yielding a perfect $P_k$. Similarly, if some reference segment is longer than the sliding window length and one window happens to be inside of a reference segment, it would yield an extra window with a perfect WinDiff.
Taking inspiration from the speaker diarization task, we propose to employ the segmentation purity and coverage metrics,
as described and implemented in \texttt{pyannote.metrics} \cite{pyannote.metrics},
for evaluation of speech topic segmentation to make it easier to compare other systems to our results. We still report the established $P_k$ and WinDiff metrics to compare systems of the same architecture.

Segmentation purity is calculated for each hypothesis segment
and shows the maximum percentage of a hypothesis segment overlapping with a single reference segment.
Segmentation coverage is calculated for each reference segment
and shows the maximum percentage of a reference segment overlapping with a single hypothesis segment.
We report the Segmentation Purity and Coverage F-score (SPCF), which is calculated as follows:
\begin{equation*}
\text{SPCF} = 2\frac{\text{Purity} \cdot \text{Coverage}}{\text{Purity} + \text{Coverage}}
\end{equation*}

\subsection{Model Optimization}
In our experimental setup, we meticulously optimized all hyperparameters for the baseline model but only with English samples, the pipeline, and end-to-end model training methodologies. We employed the Adam optimizer \cite{kingma2014adam} with an initial learning rate of 0.001. To adapt dynamically to the most appropriate learning rate, we utilized the Reduce Learning Rate on Plateau learning rate scheduler. Given the inherent variance in input lengths within our dataset, we consistently maintained a batch size of one for each training step.

The predictive task entails a binary classification problem aimed at accurately categorizing each sentence into one of two classes: \{${0,1}$\}. We apply the  Binary Cross Entropy objective function for training.

The pipeline model integrates the Insanely-Fast-Whisper\footnote{\url{https://github.com/Vaibhavs10/insanely-fast-whisper}} framework with Pyannotate-speaker-diarization-3.0 \footnote{\url{https://huggingface.co/pyannote/speaker-diarization-3.0}} to transcribe broadcasted speech into transcripts within a multi-speaker setting. These transcripts are then encoded into latent representations using SONAR's sentence encoder, where each sentence is represented by a vector of 1024 dimensions. On the other hand, the end-to-end model relies on extracting latent representations through a fixed sliding window approach. This involves using a non-overlapping window spanning 10 seconds, which traverses the entire audio recording. At each 10-second interval, the model computes the speech representation, resulting in vectors of 1024 dimensions using SONAR's speech encoder model. The encoded latent representation of both the pipeline and the end-to-end model serves as the training data for the topic segmentation model, which utilizes a transformer-based architecture with four attention heads.


\section{Results}

\subsection{Topic Segmentation Techniques}
We conducted a comparative analysis of our results with the existing traditional TextTiling approach \cite{hearst-1997-text} and the benchmark TextSeg model proposed by \citet{koshorek2018text}. As we explore the pipeline and the end-to-end approaches, we present our findings from both experiments using the English Euronews dataset. We applied the TextTiling algorithm and the TextSeg model to the transcribed English portion of the Euronews dataset and evaluated their performances accordingly. As demonstrated in Table \ref{tab:results:overall}, the pipeline system outperforms all other approaches and achieves state-of-the-art results for speech topic segmentation. The slightly lower $P_k$ score observed for the end-to-end model may be attributed to scenarios with multiple speaker overlaps, a challenge that is addressed during the transcription generation process using speaker diarization. For the end-to-end model, the SPCF estimates the trade-off between the exact prediction of the timestamp and the prediction of the window. The pipeline system is able to achieve a higher SPCF than the end-to-end one.

\begin{table}[!h]
\centering
\setlength{\tabcolsep}{0.9mm}
\begin{tabular}{l ccc}
\toprule
System & $P_k$ & WinDiff & SPCF \\
\midrule
TextTiling &0.4459 &0.4817 &0.3222 \\
TextSeg &0.3926 &0.3970 & 0.4500 \\
\textbf{Pipeline} &\textbf{0.2431}& \textbf{0.2670} &\textbf{0.6014} \\
End-to-End &0.2564 & 0.3379&0.4742 \\
\bottomrule
\end{tabular}
\caption{Results for the English Euronews dataset with different topic segmentation techniques.}
\label{tab:results:overall}
\end{table}

\subsection{Pipeline Model}

Our pipeline model investigated two distinct configurations: monolingual and multilingual. In the monolingual set of experiments, we utilize a language-specific SONAR sentence encoder, with the model head trained exclusively on the specified language of the Euronews dataset. Conversely, in the multilingual pipeline experimentation, we continue to employ the language-specific SONAR sentence encoder while training the model head across all six languages present in the Euronews dataset. The results for the monolingual pipeline model and the multilingual pipeline model can be seen in Table \ref{tab:results:pipeline}.

\textbf{Monolingual pipeline system.} The monolingual pipeline achieves state-of-the-art results for English, surpassing all other text-based topic segmentation approaches, with $P_k$ score of 0.2431, \textit{Windiff} of 0.2670, and \textit{SPCF} of 0.6014. $P_k$ remains the widely accepted metric for topic segmentation. However, we included \textit{SPCF} as a supplementary metric to provide additional insights. Our primary analysis focuses on $P_k$, consistent with its adoption by other researchers.  These metrics demonstrate our model's capability to accurately predict precise topic boundaries. Italian and French, with lower numbers of segments per recording as seen in Table \ref{tab:data:euronews}, exhibit $P_k$ scores of 0.2273 and 0.2579, respectively. This suggests that their longer segment sizes contribute to the model's ability to predict topic boundaries with high \textit{SPCF} scores. Conversely, Spanish, Portuguese, and German exhibit similar overall data durations, resulting in higher $P_k$ scores but lower \textit{SPFC}. Despite the model's capacity to predict these segments due to the higher number of segments per recording, precision slightly diminishes compared to other languages. 

\textbf{Multilingual pipeline system.} The multilingual pipeline model enhances overall performance across all six languages. We observe a gradual decrease in $P_k$ scores and \textit{Windiff} scores along with the \textit{SPCF} scores for all languages, except for Spanish. The higher number of segments per recording in Spanish leads to a sharp drop in the \textit{SPCF} score. Particularly noteworthy is the improvement for the high-resource English dataset portion with the multilingual model. 
\begin{table}[!h]
\centering
\setlength{\tabcolsep}{0.9mm}
\begin{tabular}{l ccc ccc}
\toprule
\multirow{2}{*}{Language} & \multicolumn{3}{c}{Monolingual head} & \multicolumn{3}{c}{Multilingual head}\\
\cmidrule{2-7}
 & $P_k$ & WinDiff & SPCF & $P_k$ & WinDiff & SPCF \\
\midrule
English &0.2431 &0.2670 &0.6014  &0.1988 &0.2466&0.5471 \\
French &0.2579 &0.2939 & 0.6293 &0.2406 & 0.2923&0.5362\\
German  &0.2406 &0.3100 & 0.4897 & 0.1993& 0.2536&0.4856\\
Italian &0.2273 &0.2727 & 0.5240 & 0.2184& 0.2728&0.4694\\
Portuguese &0.1968 &0.2483 & 0.4802 &0.1616 & 0.2014&0.5346\\
Spanish &0.2310 &0.2656 & 0.4536 & 0.2262&0.2800&0.3846\\
\midrule
 Average & 0.2328 & 0.2763 & 0.5297 & 0.2075 & 0.2578 & 0.4929 \\
\bottomrule
\end{tabular}
\caption{Results for the complete Euronews dataset, pipeline system.}
\label{tab:results:pipeline}
\end{table}


\subsection{End-to-End Models}

To ensure a fair comparison, we adopt a similar setup for the end-to-end model as employed in our pipeline model. Consequently, we explore two configurations: monolingual and multilingual. In the monolingual experiments, we utilize a language-specific SONAR speech encoder to train the segmentation model head exclusively on the specified language from the Euronews dataset. For the multilingual end-to-end experiments, we retain the language-specific SONAR speech encoder while training the model head across all six languages present in the Euronews dataset. We use the language-specific SONAR speech encoder for the end-to-end multilingual model. The multilingual head was trained using language-specific encoders corresponding to the language of each data sample. For example, German samples were processed using the German speech SONAR encoder. The results for both the monolingual and multilingual models are summarized in Table \ref{tab:results:end-to-end}.

\textbf{Monolingual End-to-End system.} The monolingual end-to-end model achieves competitive results for English, with a $P_k$ score of 0.2564, Windiff of 0.3379, and SPCF of 0.472. These metrics highlight the model's ability to accurately predict precise topic boundaries. Italian, Spanish, and Portuguese exhibit slightly higher $P_k$ scores compared to the pipeline model, accompanied by a decrease in SPCF score. Additionally, as these languages have lower data amounts in the Euronews dataset, the distribution trend within the data varies. Conversely, German shows overall improved results compared to the pipeline model. Despite scoring lower than the pipeline, the end-to-end speech segmentation outperforms previous text-based topic segmentation models, exhibiting lower $P_k$ and Windiff scores.

\textbf{Multilingual End-to-End system.} The multilingual end-to-end model enhances overall performance across all six languages. We observe a gradual decrease in $P_k$ scores and Windiff scores for all languages, coupled with an increase in SPCF scores for Spanish. The multilingual end-to-end results demonstrate competitive performance compared to the monolingual pipeline model.


\begin{table}[!h]

\centering
\setlength{\tabcolsep}{0.9mm}
\begin{tabular}{l ccc ccc}
\toprule
\multirow{2}{*}{Language} & \multicolumn{3}{c}{Monolingual head} & \multicolumn{3}{c}{Multilingual head}\\
\cmidrule{2-7}
 & $P_k$ & WinDiff & SPCF & $P_k$ & WinDiff & SPCF \\
\midrule
English &0.2564 &0.3379 & 0.4742 &0.2370 &0.2990&0.4799 \\
French &0.2621 & 0.2802&0.4681  & 0.2206&0.2815&0.4349 \\
German  &0.2046 &0.3367  &0.4358 &0.1784 & 0.2443&0.3928\\
Italian &0.2789 &03505  & 0.4759&0.2558 & 0.3284&0.3377 \\
Portuguese &0.2849 &0.3695 &0.5099 &0.2071 & 0.2768&0.4760 \\
Spanish &0.3313 &0.4332 &0.2796  &0.2628 &0.3390 &0.3377\\
\midrule
 Average & 0.2695 & 0.3513 & 0.4406 & 0.2270 & 0.2948 & 0.4098 \\
\bottomrule
\end{tabular}
\caption{Results for the complete Euronews dataset, End-to-End system.}
\label{tab:results:end-to-end}
\end{table}


\textbf{Effect of the encoder pretraining language.} To further assess the effectiveness of our end-to-end model in a cross-lingual setting, we conducted two distinct sets of experiments: 1) \textit{Monolingual Cross-lingual}, and 2) \textit{Multilingual Cross-lingual}.

In the monolingual model, we exclusively employed the English SONAR Speech encoder on the English Euronews train set to train the speech segmentation head. Subsequently, the English-trained head was tested across all languages from Euronews, which were produced from language-specific SONAR speech encoders. Conversely, for the multilingual end-to-end model, we applied the only English SONAR encoder to the complete Euronews train set to train the speech segmentation head. The fully-trained head was then tested across all languages from Euronews, which were produced from language-specific SONAR speech encoders. Our expectation is that the English SONAR encoder is more robust to different acoustic conditions because it is trained on a larger amount of diverse data, but these benefits might be canceled out by the language mismatch when applied to languages other than English. The results for both sets of experiments can be seen in Table \ref{tab:results:cross-lingual}.

The monolingual cross-lingual end-to-end model generally achieved lower results than the monolingual end-to-end model, except for Portuguese and Spanish, which showed competitive results with the monolingual end-to-end model. Moreover, the monolingual cross-lingual model yielded overall better results than the multilingual cross-lingual model. The primary reason for this observation is the use of only the English SONAR speech encoder with all other languages from Euronews. Additionally, English belongs to the Germanic language family, which may explain the relatively lower $P_k$ and Windiff scores observed for German.

From the monolingual cross-lingual experiments, we infer that cross-linguality manifests in the latent representation. Therefore, this model can be utilized for low-resource languages or other languages without the need for individual training.
\begin{table}[!h]

\centering
\setlength{\tabcolsep}{0.9mm}
\begin{tabular}{l ccc ccc}
\toprule
\multirow{2}{*}{Language} & \multicolumn{3}{c}{Monolingual head} & \multicolumn{3}{c}{Multilingual head}\\
\cmidrule{2-7}
 & $P_k$ & WinDiff & SPCF & $P_k$ & WinDiff & SPCF \\
\midrule
English &0.2564 &0.3379 &0.4742 &0.2845 &0.3482 & 0.4917 \\
French &0.3205 &0.3796 &0.5437 &0.3310 &0.3944 &  0.5512 \\
German  &0.3023 &0.3946 &0.4204 &0.2782 &0.3445 &  0.4823 \\
Italian &0.3162 &0.3626 &0.5978 &0.3223 &0.3703 & 0.5665 \\
Portuguese &0.2809 &0.3568 &0.5939 &0.3436 &0.3975 & 0.7213 \\
Spanish &0.3136 &0.3750 &0.4735 &0.3191 &0.3738 & 0.4454 \\
\midrule
 Average & 0.2983 & 0.3678 & 0.5173 & 0.3131 & 0.3715 & 0.5431 \\
\bottomrule
\end{tabular}
\caption{Results for the complete Euronews dataset, End-to-End system with the SONAR encoder pretrained on English data only.}
\label{tab:results:cross-lingual}
\end{table}


\textbf{Effect of domain mismatch.} To assess our model's performance in a cross-domain setting, we apply a German SONAR speech encoder and train the speech segmentation head to evaluate the results on the Tagesschau test set. We utilize solely the German-trained heads from both the Euronews and Tagesschau datasets to conduct experiments and evaluate the performance of cross-domain topic segmentation. The Tagesschau results are shown in Table \ref{tab:results:tagescchau}.

The Tagesschau-trained head exhibited a notably low $P_k$ score when tested on the Tagesschau test set. However, utilizing the German Euronews-trained head to evaluate the model's performance on Tagesschau resulted in a noticeable increase in the $P_k$ score. Furthermore, when the head was trained on both Tagesschau and Euronews datasets together, the $P_k$ score showed an improvement compared to training on Euronews only. These results indicate the domain specificity of the resulting segmentation models. The data distribution between Tagesschau and the German subset from the Euronews dataset displays fairly similar patterns, with slightly more segments per recording observed in Tagesschau, therefore we attribute the domain specificity to the differing acoustic styles of the programs.

\begin{table}[!h]

\centering
\setlength{\tabcolsep}{0.9mm}
\begin{tabular}{l ccc}
\toprule
Training data & $P_k$ & WinDiff & SPCF \\
\midrule
Tagesschau &0.1519 &0.2403 & 0.3542 \\
Euronews &0.3063 &0.3769 & 0.4767 \\
Combined &0.1859 &0.2750 & 0.3256 \\
\bottomrule
\end{tabular}
\caption{Results for the Tagesschau evaluation set, End-to-End system.}
\label{tab:results:tagescchau}
\end{table}


\textbf{Zero-shot cross-lingual evaluation.} The motivation behind conducting zero-shot evaluations in a cross-domain, cross-lingual setting is to assess the model's ability to generalize across different languages and domains. Utilizing Akashvani AIR news for this purpose offers a unique advantage as it allows for zero-shot evaluations in a language unrelated to those previously used for speech topic segmentation tasks, thereby providing valuable insights into the model's cross-lingual capabilities. Moreover, Hindi serves as a crucial testbed for exploring topic segmentation tasks due to its distinct morphological linguistic characteristics along with a wide range of phonetic characteristics.

To evaluate the performance of the end-to-end model, we present results from two different setups. We employ a Hindi SONAR encoder and evaluate the segmentation heads trained in both English monolingual and multilingual head from Table \ref{tab:results:end-to-end}. The English-trained head outperformed the multilingual counterpart, with a $P_k$ score of 0.3788 compared to 0.3938 for the multilingual head. This improvement with the English-trained head can be attributed to the presence of a few English audio segments within the Akashvani news dataset, typically speeches from international conferences. However, the competitive scores obtained by the multilingual-trained head indicate that our proposed end-to-end model is capable of achieving effective cross-lingual performance even in the context of entirely different languages, which can be seen in Table \ref{tab:results:akashvani}.

\begin{table}[!h]

\centering
\setlength{\tabcolsep}{0.9mm}
\begin{tabular}{ccc ccc}
\toprule
\multicolumn{3}{c}{English head} & \multicolumn{3}{c}{Multilingual head}\\
\cmidrule{1-6}
$P_k$ & WinDiff & SPCF & $P_k$ & WinDiff & SPCF \\
\midrule
0.3788 &0.4416 & 0.5540 & 0.3938 &0.4695 & 0.4810 \\
\bottomrule
\end{tabular}
\caption{Results for the Akashvani evaluation set, End-to-End system.}
\label{tab:results:akashvani}
\end{table}


\section{Conclusion and Future Work}
In this work, we introduce a new large speech topic segmentation dataset based on real-world publicly available data, which provides training, validation, and evaluation sets in multiple languages from multiple sources, and evaluate the usage of pretrained multilingual multimodal semantic embeddings on this dataset.

The proposed dataset comprises over 1000 hours of recordings from three sources covering seven languages in total: English, French, German, Italian, Portuguese, Spanish, and Hindi. This diverse and realistic data is accompanied by high-quality labels. Topic segments are annotated by a native speaker for the Hindi recordings, and the recordings in other languages are segmented by the content publishers. German recordings are represented by two sources to evaluate the domain robustness of speech topic segmentation approaches. Hindi recordings from another source are included in the evaluation set only and serve as a showcasing opportunity for both cross-domain and zero-shot multilingual capabilities of speech topic segmentation systems. The original audio is publicly available, and this makes it easy for anyone to experiment with innovative systems of any architecture.

We design a pipeline system comprising the off-the-shelf speaker diarization and ASR systems coupled with one of three text topic segmentation models. Two of the investigated topic segmentation methods are the traditional non-neural TextTiling algorithm and the well-known TextSig neural network model. The third method is developed by us and improves TextSeg with Transformer architecture instead of BLSTM, stronger pretrained semantic encoder, tighter context integration, and more efficient output formulation. Our method outperforms the other two for English with a $P_k$ score of 0.2431, compared to 0.3926 for TextSeg and 0.4459 for TextTiling. The multilinguality of the utilized sentence encoder allows simple scaling up of the training to multiple languages, further improving the $P_k$ score of our system to 0.1988 for English. Furthermore, our multilingual system performs reasonably well in another six languages, including one not present in the training data. To the best of our knowledge, we present the first speech topic segmentation systems for such languages as German, Italian, Portuguese, Spanish, and Hindi. Finally, we explore the multimodal properties of the utilized semantic encoder and transform our system to an end-to-end model for the direct topic segmentation of speech without relying on ASR preprocessing. While its performance is worse compared to the pipeline approach, it can offer better computational efficiency and has the potential for incorporation of additional cues, which appear in speech only.

Future work should focus primarily on advancements of the end-to-end approach to reach and outperform the pipeline results. A deeper adaptation of the speech encoder to the task can be achieved through parameter-efficient finetuning and would allow the extraction of more relevant features from input speech. Overlapping input segments or ensembling with a speaker diarization model can be utilized to improve the time precision of the end-to-end model. Our experiments uncover the limited cross-domain generalization, and this requires additional investigation, potentially involving data augmentation. More training data can also be added with further scaling of multilingual training, and extension to multimodal training, as text data can be found in larger quantities. Another promising direction of transfer learning is the inclusion of topic classification tasks, which can both improve the segmentation results and add more use cases for the model by making its outputs more interpretable. Finally, topic clustering methods for text,
 such as BERTopic \cite{grootendorst2022bertopic}, might be transferrable to speech modality as well.






\bibliography{mybibfile}

\end{document}